\documentclass[a4paper]{article}
\usepackage{spconf,amsmath,graphicx,cite}
\usepackage{graphicx}
\usepackage{amsmath}
\usepackage{amssymb}
\usepackage{hyperref}
\usepackage{placeins}

\usepackage[table,x11names]{xcolor}
\usepackage{amssymb}
\usepackage{amsbsy}
\usepackage{amsthm, mathrsfs}
\usepackage{mathtools}
\usepackage{array} 
\usepackage{cite}
\usepackage{dsfont}
\usepackage{subcaption}
\usepackage{caption}
\usepackage{graphicx}
\usepackage{caption}

\theoremstyle{definition}
\title{Privacy-Preserving Image Sharing via \\Sparsifying Layers on Convolutional Groups}
\name{Sohrab~Ferdowsi$^*$, Behrooz~Razeghi$^*  \, \! ^{\ddagger}$, Taras~Holotyak$^*$, Flavio P. Calmon$^\dagger$, Slava~Voloshynovskiy$^*$%
	\thanks{$^{\ddagger}$ Work done while at Harvard University. B.\,Razeghi has been supported by the ERA-Net project
    ID\_IoT No 20CH21\_167534.}
    \thanks{Implementation codes available at: \scriptsize{\url{https://github.com/sssohrab/sparsifying_groups_imAmbiguation}}}
}
\address{%
    \tabular{c}
		$^*$ University of Geneva
	\endtabular
	\hskip 0.5in
    \tabular{c}
		$^\dagger$ Harvard University
	\endtabular
}

\begin{document}

\maketitle
%
%
%%%%%%%%%%%%%%%%%%%%%%%%%%%%%%%%%%%%%%%%%%%%%%%%%%%% 
%%%%%%%%%%%%%%%%%%%%%%
%%%%%%%%%%%%%%%%%%%%%%        Abstract
%%%%%%%%%%%%%%%%%%%%%%
%%%%%%%%%%%%%%%%%%%%%%%%%%%%%%%%%%%%%%%%%%%%%%%%%%%%
%
%
\begin{abstract}
We propose a practical framework to address the problem of privacy-aware image sharing in large-scale setups. We argue that, while compactness is always desired at scale, this need is more severe when trying to furthermore protect the privacy-sensitive content. We therefore encode images, such that, from one hand, representations are stored in the public domain without paying the huge cost of privacy protection, but ambiguated and hence leaking no discernible content from the images, unless a combinatorially-expensive guessing mechanism is available for the attacker. From the other hand, authorized users are provided with very compact keys that can easily be kept secure. This can be used to disambiguate and reconstruct faithfully the corresponding access-granted images. We achieve this with a convolutional autoencoder of our design, where feature maps are passed independently through sparsifying transformations, providing multiple compact codes, each responsible for reconstructing different attributes of the image. The framework is tested on a large-scale database of images with public implementation available.
\end{abstract}


%%%%%%%%%%%%%%%%%%%%%%%%%%%%%%%%%%%%%%%%%%%%%%%%%%%% 
%%%%%%%%%%%%%%%%%%%%%%
%%%%%%%%%%%%%%%%%%%%%%     Introduction
%%%%%%%%%%%%%%%%%%%%%%
%%%%%%%%%%%%%%%%%%%%%%%%%%%%%%%%%%%%%%%%%%%%%%%%%%%%
%
%
\section{Introduction}

\vspace{-8pt}

%%      First Paragraph: Motivation
%%      What’s the problem/opportunity?
%%
%%
%\textcolor{red}{Behrooz, as suggested by Flavio, I am changing everywhere the term ``data release'' with ``data sharing''. Please change every instance you see.}
%% Okay Sohrab jan

%% Sohrab jan, I think the second setence below is better. 
% A prime challenge in data science is how to reconcile, from one hand, the need to provide \textit{useful representation} of data as is required by the numerous machine learning tasks, and from the other hand, the growing concern about the \textit{privacy constraints} of individuals and the associated data, or content anonymity of proprietary data when sharing with third parties.
In the era of big data, with recent advances in data driven machine learning frameworks coupled with growing concern about the privacy of individuals' identities %and their associated data 
when shared with third parties, it is desirable to release \textit{useful representation} of data while simultaneously satisfying some \textit{privacy constraints}. 
We consider scenarios where the data owner collects massive amounts of data to provide some utility for the authorized users (clients). For instance, governments, social media or fin-tech databases posses facial images of citizens or customers that should be efficiently communicated with several of their trusted partners, while strictly having to protect the privacy of the individuals.
%, or e.g., the social networks may have to share user's data for various processes with different data centers. 
%The massive amounts of data, along with the variety of needed tasks, naturally requires the data to be shared within different servers
% [A problem with SKA: No strict need for privacy] For instance, the Square Kilometre Array Organisation (SKAO) \cite{} provides high resolution images for astronomers, each being about 4GB in size, or a ... .
%Due to the massive amount of data, storage and computational capacities, the data owners outsource their data collection to third parties (servers). 
The engineering goal in such cases is to design a practical multi-party data-sharing mechanism with constraints on \textit{data utility} and \textit{data privacy}.

%% Second Paragraph: Our setup
%--------------
%   Our work
%--------------

% %group representative based schemes \cite{gheisari2019aggregation}, Sparse Coding with Ambiguation (SCA) \cite{Razeghi2017wif}, 
%%         
%%. abadi2009data,
%% helper Data Systems (HDS) \cite{LT2003}
There is a rich literature of prior work on privacy-assuring mechanisms based, e.g., on cryptographic methods \cite{aguilar2013recent}, differential private based techniques \cite{dwork2008differential}, generative adversarial models \cite{huang2017context, tripathy2017privacy} or embedding based schemes \cite{gheisari2019aggregation}, just to name a few. 
The classical privacy-preserving image release mechanisms were mostly based on obfuscation techniques, such as pixelization and bluring \cite{hill2016effectiveness, blog2012face}, or partial encryption such as P3 \cite{ra2013p3}, which are defeated using machine learning methods \cite{mcpherson2016defeating}. 
%Among the more recent methods, the privacy-preserving image sharing 
%
%adversarial model addresses the privacy-utility trade-off as a constrained min-max game between a defender and an adversary \cite{huang2017context, tripathy2017privacy}.% huang2018generative
The framework of Sparse Coding with Ambiguation (SCA) \cite{Razeghi2017wifs, Razeghi2019icip} shares a compressed, but ambiguated representation of data within the public domain, while the users can benefit some utility given an authorized query. 
%Our work is closely related to the Sparse Coding with Ambiguation (SCA)  privacy-preserving framework \cite{Razeghi2018icassp, Razeghi2019icip}. 
This work is closely related to the SCA, however, rather than information-theoretic arguments, our focus here is mostly computational: 
Firstly, we formulate the requirements of a privacy-aware data-driven image sharing mechanism, where the data is split between public and trusted parties. 
Secondly, we provide an actual implementation of this framework in practice for the case of images using Convolutional Neural Networks (CNNs). 
% 
% 
%Consider a three-party data release scenario involves (a) a data owner, (b) data users, and (c) service provider(s). 
%The data owner outsources some representations of the image data that he owns to `honest but curious' server(s). 
%The data owner attempts to: (1) protects original data collection from server side analysis; (2) provides a pre-determined utility for his authorized clients; (3) protects original data collection against the un-authorized parties. 
%We explicitly define our measure of utility and privacy as capability of reconstruction for authorized and un-authorized parties, respectively.  
%Moreover, in order to follow Kerckhoffs's Principle in cryptography, we assume that the data-release mechanism is publicly known. 
%
%\textcolor{red}{[This paragraph should also be changed, since we are no longer talking about identification.]}
%\textcolor{blue}{Are you sure? Since here we talk our setup for general Outsource services. }

Concretely, consider a three-party image/visual information sharing or usage scenario that involves (a) a data owner, (b) data users, and (c) service provider(s). The data owner outsources some representations of the images that s/he owns to the `honest but curious' server(s) for storage or further communication or sharing with data users. 
S/he attempts to: (1) protect the original data collection from server side analyses interested in knowing the data content provided by the data owner; (2) provide a pre-determined utility (e.g. reconstruction) for his/her authorized clients; (3) protect original data collection against the un-authorized parties. In order to follow Kerckhoffs's Principle in cryptography, we further assume that the data-sharing mechanism is publicly known, but a secret key used for the data protection is kept secret. 

To avoid expensive solutions to provide security for privacy-sensitive data, e.g., through cryptographic encryption, we propose to ambiguate the data representation (as detailed next), and instead provide security only for the disambiguation key. This scheme, of course, is only useful when the key is much smaller in size than the original data\footnote{So e.g., a na\"{i}ve permutation of image pixels is not useful, since the permutation map is even larger than the original image.} (before and after ambiguation), for which we provide a practical solution. A key justification for this double-splitting of the data is the fact that compactness, while always desired at large-scale setups, is much more of a cruicial need for security and privacy solutions, both storage and communication. 

% Change this paragraph.
%A core motivation for our setup is the \textit{fundamental trade-off} between the amount to which we can provide secure communication and storage, from one hand, and the entropy of the data, from the other hand. While in large-scale setups, compression is always an issue, e.g., for fast communication, ensuring the security of the data (both from information-theoretic and computational perspectives), makes this need much more severe. Therefore, it is indeed desired to provide a double-split, such that the higher-entropic portion is not secured, but is ambiguated, and only the (much) lower-entropic part corresponding to the a compressed data representation is secured. This is much cheaper than securing the original data. We therefore, define our measure of utility and privacy as the quality and the computational resources needed for reconstruction of the data for the authorized and un-authorized parties, respectively, and as functions of the compression levels of the representations. 

%\textcolor{red}{Behrooz: Sohrab, I read and revised from beginning till here. Now, I jump to the Section 2. Please revise the following.}
While there is an extensive literature on using traditional image compression codecs to provide security along with compression, (e.g., see \cite{8537968, 7178165, 7025694, 1421853}), an important limitation arises with these approaches when they are used in large-scales and possibly for domain-specific images. Since in such scenarios images have similar encoding (e.g., high concentration of activation of DCT coefficients at certain regions), the adversary can benefit from this to infer the statistics of encoded images.\footnote{However, if the image compression bases and quantizers are learned, the network tries to spread-out the activities of the representations, as this provides better rate-distortion trade-offs for the learned network.} Moreover, even the compression capability of standard codecs have been seriously challenged by learning-based solutions. This has led to an active area of research that uses deep learning (see e.g., \cite{balle2016end, agustsson2018generative}) to learn optimal compression. This work follows the learning-based approach, however, instead of the usual binary representations used in these methods, we focus on sparsity of representations similarly to \cite{Sohrab_Thesis}. This allows us to benefit from the SCA framework  \cite{Razeghi2018icassp, Razeghi2019icip} for ambiguation.% Therefore, we come up with a network capable of producing sparse codes for very large images, a major challenge addressed in this paper.

This paper presents two main contributions: Firstly, we introduce a data sharing setup as detailed in section \ref{sec:architecture}, where the cost of security is minimized in terms of the amount of bits required, thanks to our end-to-end solution for capturing data redundancies with representation learning. 
%We then discuss the criteria for this setup to be useful in terms of the required computational complexities, the storage cost and the reconstruction fidelity. 
Secondly, we provide a practical and scalable solution with two particular architectural novelties for CNNs: 1) Multiple code-maps using fully-connected groups on convolutional filters. 2) The $k$-sparsity non-linearity in CNNs along with ReLUs without slowing down training. This is detailed in section \ref{sec:architecture}. The experimental setup and concluding remarks are then presented in sections \ref{sec:exp} and \ref{sec:conclusions}, respectively.

\vspace{-10pt}

\section{Privacy-preserving Image Sharing}
\label{Sec:Privacy-preservingImageRelease}

\vspace{-7pt}

We encode the data $\mathbf{x} \in \mathcal{X}$, as the pair $(\mathbf{u}_p, \mathbf{u}_s)$, where $\mathbf{u}_p \in \mathcal{U}_p$ is the part of the data stored in public, while $\mathbf{u}_s \in \mathcal{U}_s$ is secured and is kept private. We require that, firstly, the representation size of $\mathbf{u}_s$ in bits be much smaller than that of $\mathbf{u}_p$. Secondly, the guessing cost of the data only given the public portion, i.e., $\mathbf{x} | \mathbf{u}_s$ should be exponential, while its reconstruction provided both parts, i.e., $\mathbf{x} | \mathbf{u}_s, \mathbf{u}_p$ should be linear.

%figure---------------------------------------------------------------------------------------
 \begin{figure}[!t]
   \begin{center} 
\includegraphics[width=0.46\textwidth]{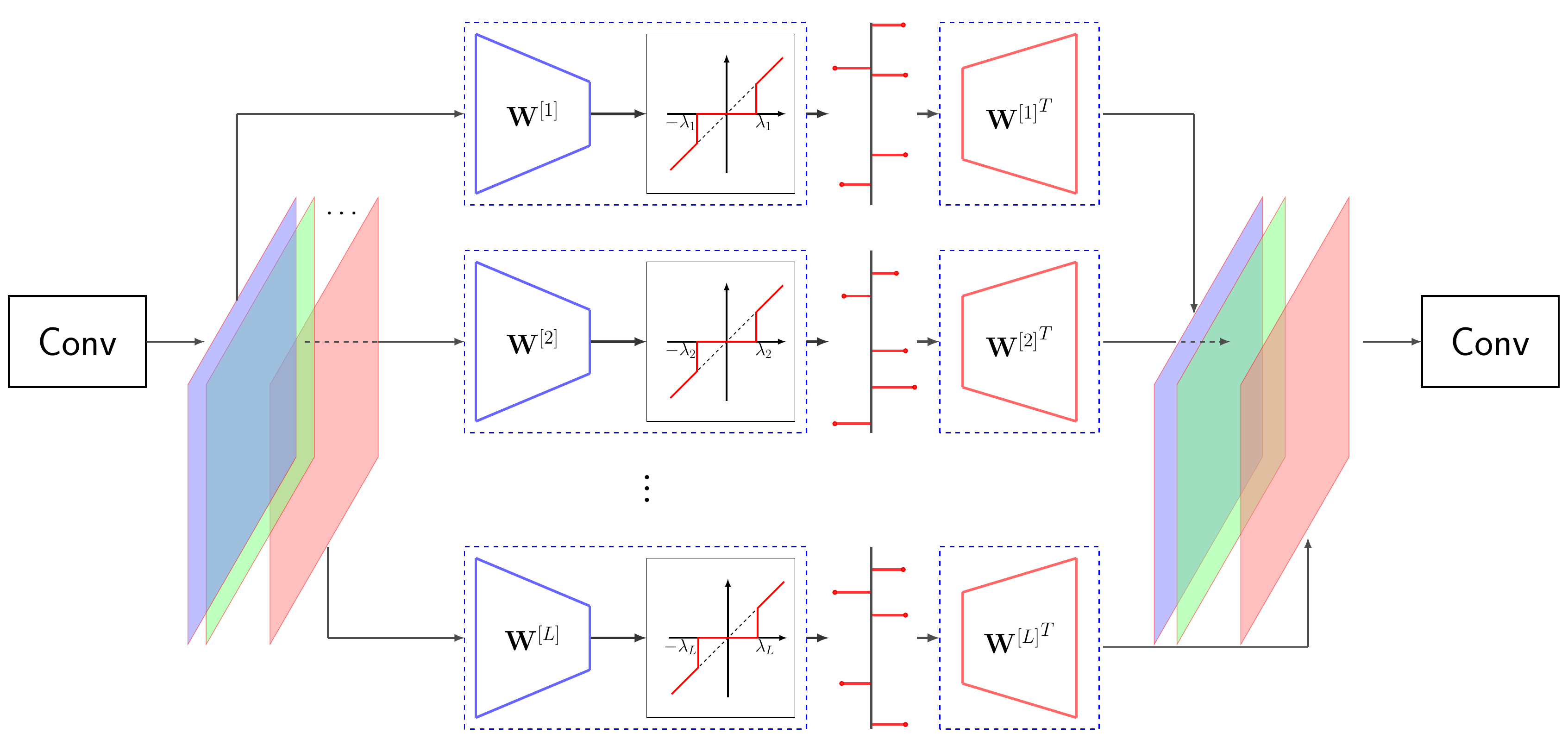}
\end{center}
\vspace{-13pt}  
   \caption{The grouped linear block to produce sparse code-maps: In the encoder side, convolutional feature maps are independently fed to fully-connected linear layers and sparsified. Symmetrically, the decoder uses tied connections and reconstruct the convolutional feature maps. This block is put in the middle of the network.}
   \vspace{-9pt}
   \label{fig:linear_layer}
   \end{figure}
%---------------------------------------------------------------------------------------------

%%%%%%%%%%%%%%%%%%%%%%%%%%%%%%%%%%%%%%%%%%%%%%%%%%%% 
%%%%%%%%%%%%%%%%%%%%%%     Scheme Overview
%
%
\vspace{-10pt}
\subsection{Proposed Scheme Overview}\label{subsec:SchemeOverview}
\vspace{-5pt}
%
%\textcolor{red}{[To Behrooz: Are we talking about identification problem here? I guess we'd better not for this paper, because it becomes more difficult and we do not have experimental support for that. So maybe we stick only to secure communication for now. In this case, is it still a 3-party setup? Do we distinguish between server and data user? For identification problem I see how, but for here I don't.]}

We achieve the above constraints with the following steps:% (Fig.~\ref{Fig:Model}):

\vspace{-1pt}

%%.   Sohrab inja ro eslah kon ba asas  framework 
%\noindent
%%
%1) \textit{Preparation at Owner Side:}
%The owner generates the sparse codewords from the data that he owns using the  
%\textit{network-trained sparsifying transforms}. 
%Next, he shares the privacy-protected sparse codebook with the service provider (server). 
%%Following Kerckchoffs's Principle in cryptography, the data owner makes the learned sparsifying transform publicly available. 
%%
%\noindent
%%
%2) \textit{Querying at Data User Side:}
%The data user generates a sparse representation from his query data using the  \textit{shared sparsifying transform}. 
%Then, the client sends a function of his sparse representation to the server. 
%%
%\noindent
%%
%3) \textit{Regenerating at Server Side} 
%Given the requested probe, 
%the server runs a near neighbor search to find the stored sparse codes that are most similar (close) to the probe. 
%%
%Finally, based on the pre-determined service to the data users, the server sends back an answer to the data user. 

1) \textit{Neural network training:} A network is first trained on a sub-collection of the data, such that it produces compact sparse codes. This is shared with the public domain.
\noindent 

2) \textit{Data owner encoding and ambiguating:} The data owner uses the trained network to sparsely encode all images s/he possesses. The support of these codes are then kept secure (e.g., through encryption or secure communication), and shared with their corresponding authorized parties. The collection of all sparse codes are then ambiguated, as will be detailed below, and then shared with the public domain.
\noindent 

3) \textit{Data users/parties disambiguating their content:} The individuals or trusted parties are provided with the indices of the images that they have access to, as well as the secured supports. This acts as the key to unlock their access-granted content from the public ambiguated database.

%  \begin{figure}[t!]
%     \centering
%         \begin{subfigure}[h]{0.23\textwidth}
%         \includegraphics[scale=0.269]{DownConvBlock.pdf}%
%       % \vspace{-5pt}
%          \caption{ }
%           \label{fig:DownConvBlock}
%       \end{subfigure}
%           %
%          \hspace{-5pt} ~ \hspace{-5pt}
% %%%
%       \begin{subfigure}[h]{0.23\textwidth}
%       \includegraphics[scale=0.269]{UpConvBlock.pdf}%
%       % \vspace{-5pt}
%         \caption{}
% %         %  \vspace{-10pt}
%         \label{fig:UpConvBlock}
%     \end{subfigure}%
% %%%%
%      \caption{Convolutional blocks for down- and up-scaling.}
%   \vspace{-14pt}
%     \label{Fig:tSNE}
% \end{figure}
%

%%%%%%%%%%%%%%%%%%%%%%%%%%%%%%%%%%%%%%%%%%%%
%%%%%%%%%%%%%%%%%%%%%%
%%%%%%%%%%%%%%%%%%%%%%     Auto-enc + SCA
%%%%%%%%%%%%%%%%%%%%%%
%%%%%%%%%%%%%%%%%%%%%%%%%%%%%%%%%%%%%%%%%%%%

\vspace{-10pt}

%\subsection{Autoencoders Producing Ambiguated Codes} 
\subsection{Ambiguated Sparse Code Generation} \label{subsec:Autoencoders_Ambiguation}

\vspace{-5pt}

\paragraph*{Training Phase.} 
Given the data samples $\mathbf{x} \in \mathbb{R}^n$, 
we train a bottlenecked auto-encoder structure consisting of $L$ independent encoders, $\mathrm{Enc}^{[1]}(\cdot), \cdots, \mathrm{Enc}^{[L]}(\cdot)$, where input data variable $\mathbf{x}$ is encoded to $L$ new representations as $\mathbf{z}^{[l]} = \mathrm{Enc}^{[l]}(\mathbf{x}),\forall l \in [L]$ with $\mathbf{z}^{[l]} \in \mathbb{R}^m$, and is (approximately) reconstructed as $\hat{\mathbf{x}}^{[l]} = \mathrm{Dec}^{[l]}[\mathbf{z}^{[l]}]$, $\forall l \in \left[L\right]$. 
Furthermore, the encoding is performed such that the codes are $k$-sparse, i.e.,  $\mathrm{card} \left( \mathrm{supp}\left( \mathbf{z}^{[l]} \right)\right) = k, \forall \, l$, where $\mathrm{supp} \left( \mathbf{z}^{[l]} \right)$ is index set of nonzero entries of $\mathbf{z}^{[l]}$ and $\mathrm{card} \left( \cdot \right)$ denotes cardinality of the set. 
The sparsity level $k$ controls the reconstruction fidelity of our auto-encoding mechanism.

\vspace{-5pt}

\paragraph*{Sharing Phase.} 
The generated sparse representations are then ambiguated and shared to the public service provider. 
Given the sparse representation $\mathbf{z}^{[l]}$ with sparsity level $k$, the ambiguation mechanism $A \left( \cdot \right)$ adds $k_n \geq 0$ random noise components to the orthogonal complement of $\mathbf{z}^{[l]}$, with the same statistics as sparse code to guarantee the indistinguishably in the statistical properties. 
Therefore, we have:\vspace{-6pt}%
\begin{eqnarray}\label{Eq:Ambiguization}
\mathbf{u}_p^{[l]}   = A \left( \mathbf{z}^{[l]}  \right) = \mathbf{z}^{[l]}  \oplus \mathbf{n}_{\mathrm{supp}}, 
\end{eqnarray}
where $\mathbf{n}_{\mathrm{supp}}$ is random ambiguation noise added to the support complement of $\mathbf{z}^{[l]}$ and $\oplus$ denotes the direct sum. Note that ${\Vert \mathbf{u}_p^{[l]} \Vert}_0 = k + k_n = k'$. 
The  ambiguated sparse representations $\mathbf{u}_p^{[l]}, \forall \, l $ are then shared to the public domain. 
The support of the latent representation $\mathbf{z}^{[l]}$, denoted by $\mathbf{u}_s^{[l]} = \text{supp}\left( \mathbf{z}^{[l]} \right)$, is considered as the secure part of our data, which is shared with the private data users. 
%This secure part can even be encrypted with low complexity. 

\vspace{-5pt}

\paragraph*{Reconstruction Phase.} 
Given the support information $\mathbf{u}_s^{[l]}, \forall l \in [L]$, the service provider can reconstruct the data as: $\hat{\mathbf{x}}^{[l]} = \mathrm{Dec}^{[l]}  ( \mathbf{z}^{[l]} \mid \mathbf{u}_s^{[l]} )$. 
Our encoding introduces a concept of \textit{shared secrecy} based on support intersection of latent representations. 
We consider two hypotheses for support secrecy. 
$\mathcal{H}_1$: The authorized support secrecy $\mathbf{u}_s$, 
$\mathcal{H}_0$: The un-authorized support, generated and claimed by an adversary.

%While the secure part of our decomposition is simply considered to be $\mathbf{u}_s^{[l]} = \text{supp}\left( \mathbf{z}^{[l]} \right)$, the public part is ambiguated with additive noise as suggested in SCA mechanism \eqref{Eq:Ambiguation}, i.e., $\mathbf{u}_p^{[l]} = \mathbf{z}^{[l]} \oplus \mathbf{n}_{\mathrm{supp}}$, where $\mathbf{n}_{\mathrm{supp}}$ is imposed ambiguation noise to the \textit{non-informative} components of $\mathbf{z}^{[l]}$, and its values have the same statistics with the code. Note that ${\Vert \mathbf{n}_{\mathrm{supp}} \Vert}_0 = k' -  k \geq 0$. 

%% Behrooz :  Sohrab, I reshape the section 2.3 completely, till here. 
% tomorrow I will be on this paper with you :)

As a result of this encoding, the number of bits required to store an item of the secure part is: 
\begin{equation}
H(\mathbf{u}_s) = \log_2{{m\choose k}^L} \simeq m L \times H_2 \Big(\frac{k}{m} \Big),
\end{equation}
where $H_2(\alpha) = -\alpha \log_2{\alpha} - (1 - \alpha) \log_2{(1 - \alpha)}$ is the binary entropy, and the approximation follows the Stirling's.
From the other hand, $\mathbf{u}_p$ is ambiguated and is $k'$-sparse, with $k' \geqslant k$ and for a typical 32-bit quantization of the non-zero values, requires approximately  $H(\mathbf{u}_p) \simeq 32 m L \times H_2 \big( \frac{k'}{m} \big)$ bits of storage per item. 
The adversary should then make ${k'\choose k}^L$ guesses to reconstruct each item, i.e., $\mathcal{O}\big( \exp{H(\mathbf{u}_s)} \big)$. 
%%%%%%%%% 

% %figure---------------------------------------------------------------------------------------
%  \begin{figure}  [!t]
%   \begin{center} 
% \includegraphics[width=0.47\textwidth]{EncoderNetwork.pdf}
% \end{center}
% %\vspace{-.75cm}  
%   \caption{Encoder network. The structure of decoder is symmetric.}
%   \label{fig:Encoder}
%   \end{figure}
% %---------------------------------------------------------------------------------------------

%%%%%%%%%%%%%%%%%%%%%%%%%%%%%%%%%%%%%%%%%%%%%%%%%%%% 
%%%%%%%%%%%%%%%%%%%%%%        Architecture
%%%%%%%%%%%%%%%%%%%%%%%%%%%%%%%%%%%%%%%%%%%%%%%%%%%% 
\vspace{-5pt}

\section{Autoencoder Architecture}\label{sec:architecture}
\vspace{-5pt}
We need a practical autoencoder architecture that has three properties: Firstly, it should be bottlenecked to provide compact codes. This excludes some of the famous neural regressors like the U-net \cite{ronneberger2015u}, since they are not bottlenecked because of their skip-connections directly from the input to the output. Secondly, we need sparsified codes. The usage of sparsity-inducing non-linearities, however, is rare in the deep learning literature. The few examples available correspond to the line of work of ``unrolling iterative algorithms as neural networks'', e.g., as in LISTA \cite{gregor2010learning, xin2016maximal}, where sparsifying operators like the soft- or hard-thresholding functions are used as non-linearities within neural networks. This, however, is not common within the representation learning community. 

Thirdly, we should be able to reconstruct images with arbitrarily chosen levels of fidelity, corresponding to a prescribed representation budget for $\mathbf{u}_s$ (and also a practical upper-bound for representation of $\mathbf{u}_p$). %This suggests multiple representations, instead of one, where each code is taking care about some aspect of the image. 

To satisfy these requirements, we propose the ``sparsifying linear layers on groups'', as is schemed in Fig. \ref{fig:linear_layer}. Note that, while the fully-connected linear layers in the literature are used in CNNs, mostly to bridge the convolutional features with the one-hot encoding of softmax for classification, we use them to diversify the activities of codes, since convolutional features are highly correlated and most activities are concentrated on small sub-sets of the feature space.

Note that it is not practical to simply reshape all convolutional filters and feed them directly to a linear layer, since this would require an extremely large matrix with intractable complexity and a very high risk of over-fitting. Therefore, we take each convolutional feature separately and pass it to a much smaller fully-connected linear layer, as if we have a large matrix with block-diagonal sparsity. 
%This, furthermore, serves our requirement of having multiple codes. 
As a byproduct, we notice that the network learns multiple codes, each describing different attributes of the image.
% I may add this later.
%As a useful byproduct, we notice that after training the network, each code turns out to be responsible for one different aspect of the image, sometimes providing semantic disentanglement. This, has also been confirmed for other applications in deep learning, where the use of grouped convolutions encourages such disentanglement. 

As far as the sparsifying operator is concerned, we craft a custom non-linearity (introduced in \cite{Sohrab_Thesis}) that only passes the $k$ elements with largest magnitude, and zeros out other coefficients. Note that this, in fact, is an adaptive version of the hard-thresholding function, where the threshold is adapted to each input sample to choose only $k$ elements out of $m$.

%%figure--------------------------------------------------------------------------------%-------
% \begin{figure}  [!t]
%   \begin{center} 
%\includegraphics[width=0.45\textwidth]{downConvBlock.png}
%\includegraphics[width=0.45\textwidth]{upConvBlock.png}
%\end{center}
%\vspace{-.75cm}  
%   \caption{Convolutional blocks for down- and up-scaling.}
%   \label{fig:ConvBlocks}
%   \end{figure}
%---------------------------------------------------------------------------------------------

\vspace{-9pt}

%%%%-------   Page 4

 \begin{figure*}[!h]
    \centering
    \hspace{-15pt}
        \begin{subfigure}[h]{0.23\textwidth}
        \includegraphics[width=\textwidth]{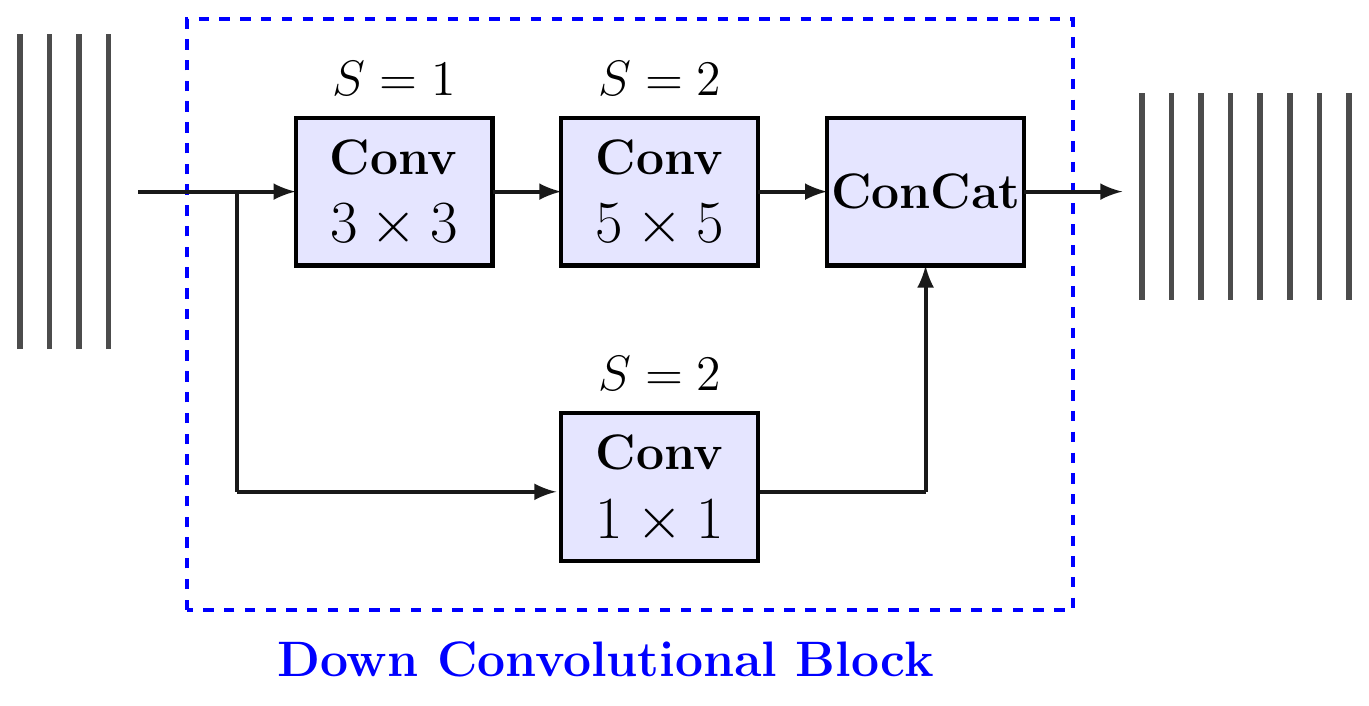}%
        \vspace{-5pt}
         \caption{ }
          \label{fig:1}
      \end{subfigure}
         \hspace{2pt} ~  \hspace{2pt}
 % \hspace{-10pt}~ \hspace{-10pt}
%%%
       \begin{subfigure}[h]{0.27\textwidth}
       \includegraphics[width=\textwidth]{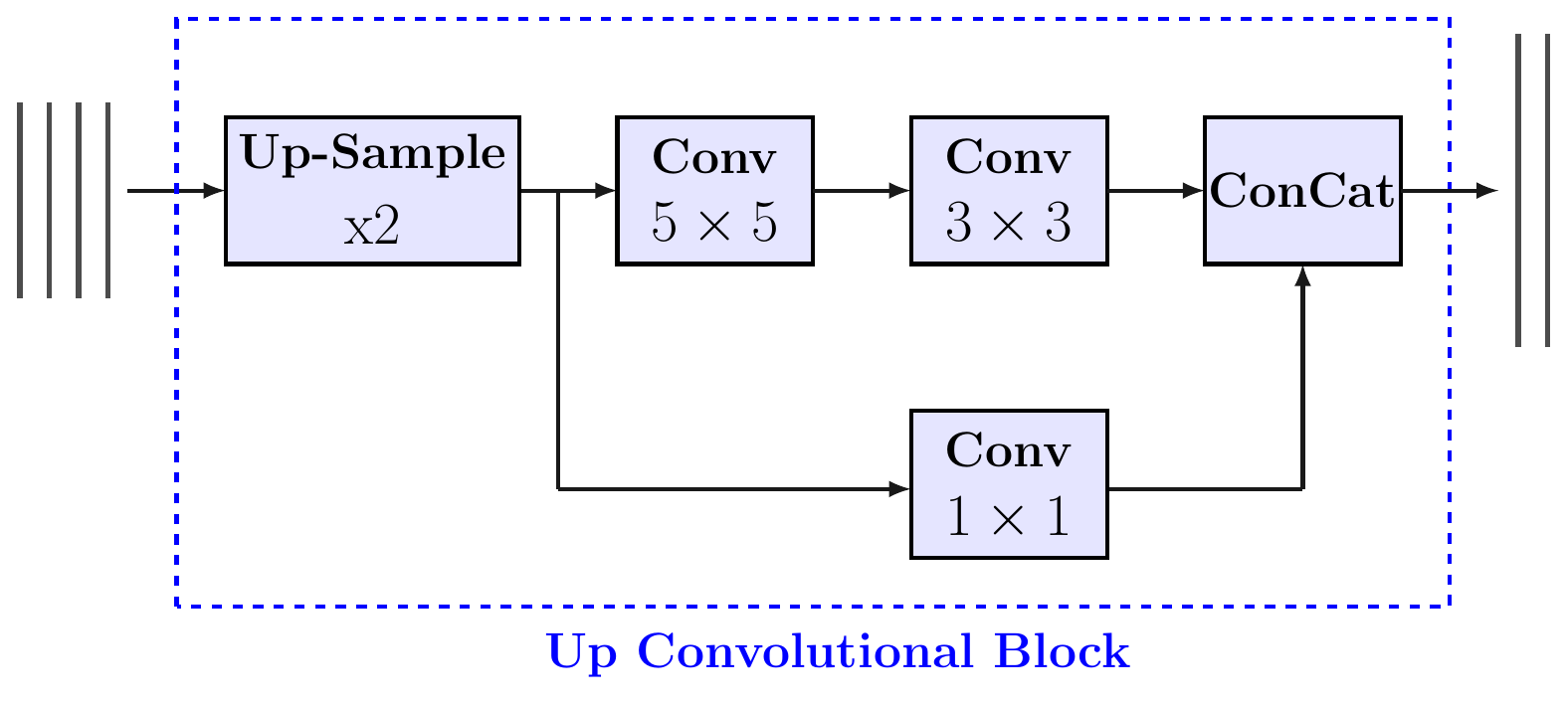}%
       \vspace{-5pt}
        \caption{}
%         %  \vspace{-10pt}
        \label{fig:2}
    \end{subfigure}
    \hspace{2pt} ~ \hspace{3pt}
  %\hspace{-10pt}~ \hspace{-10pt}
%%%%
      \begin{subfigure}[h]{0.4\textwidth}
       \includegraphics[width=\textwidth]{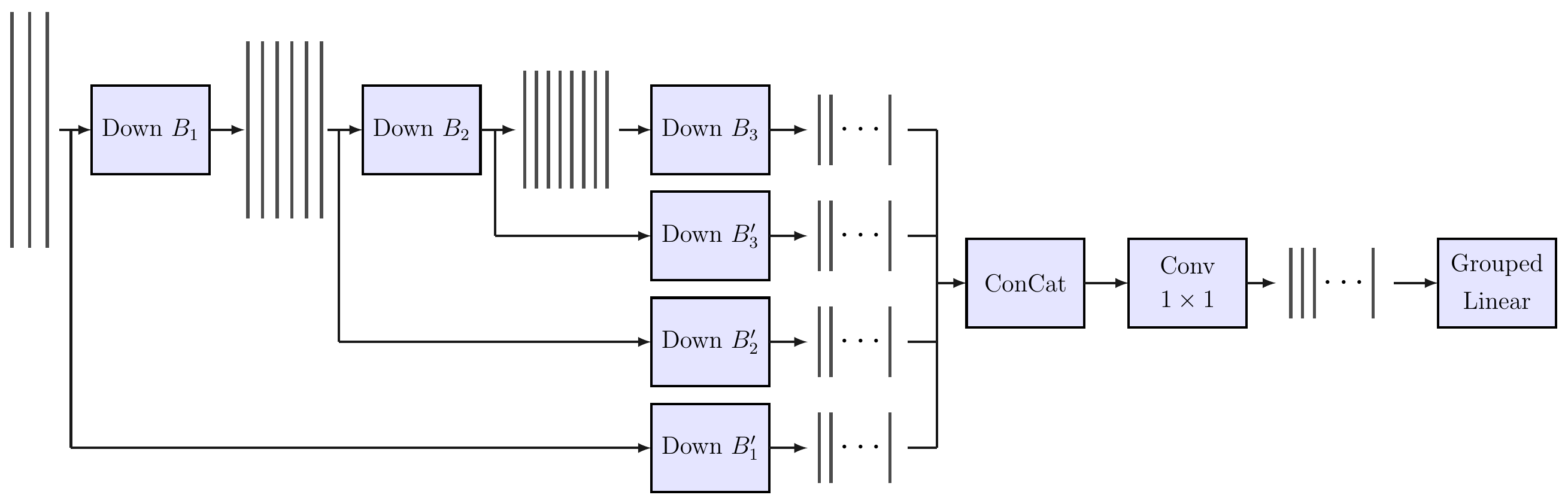}%
        \vspace{-5pt}
        \caption{}
%         %  \vspace{-10pt}
        \label{fig:3}
    \end{subfigure}
    \vspace{-9pt}
   % \caption{Visualizations of seven-class synthetic data using t-SNE. a) original domain, b) transform domain with 50\% ambiguation level, c) transform domain with 75\% ambiguation level.}
     \caption{(a) Convolutional down-sampling block. (b) Convolutional up-sampling block with bilinear interpolation. (c) Structure of the encoder of the network, where $B_i$ blocks are according to (a), and $B'_i$ blocks use addition instead of concatenation for skip-connections. The Grouped Linear block is sketched in Fig. \ref{fig:linear_layer}. Note that the decoder network is symmetrical to the encoder. }
  %\vspace{-14pt}
    \label{Fig:Architecture}
\end{figure*}

% \begin{figure*}[!t]
% \begin{center} 
% \includegraphics[scale=0.27]{DownConvBlock.pdf}%
% ~\hspace{5pt}~
% \includegraphics[scale=0.27]{UpConvBlock.pdf}%
% ~\hspace{5pt}~
% \includegraphics[scale=0.27]{EncoderNetwork.pdf}
% \end{center}
% \end{figure*}

% Sohrab, I set this figure to be located at the end of Page 4 
%%. when you add captions, the references will go to new page. 
%
% Thank you Behrooz, now it's better. I still need to remove half a page =(
%
% Yes, now even we have more space for detailed captions
%
%figure---------------------------------------------------------------------------------------
 \begin{figure*}  [!h] % or [!t]
   \begin{center} 
\includegraphics[width=0.83\textwidth]{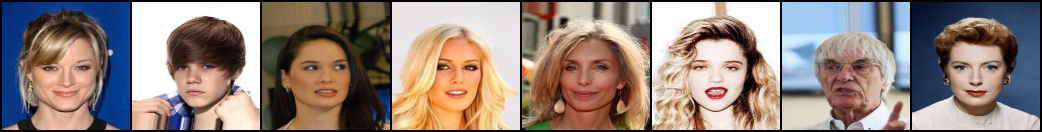}
\includegraphics[width=0.83\textwidth]{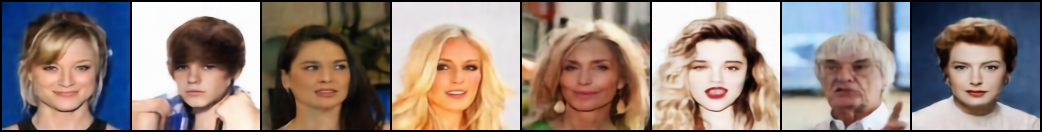}
\includegraphics[width=0.83\textwidth]{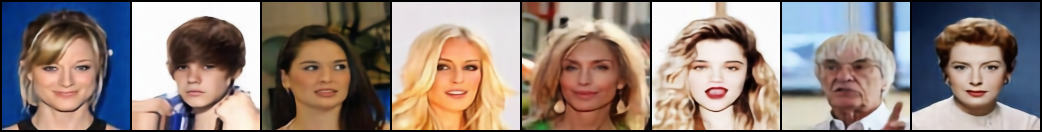}
\includegraphics[width=0.83\textwidth]{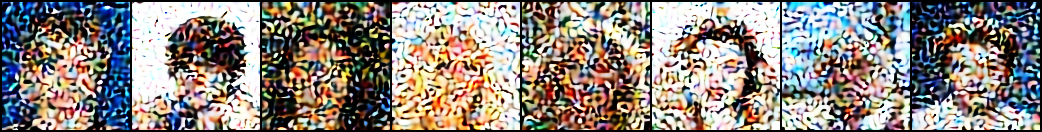}
\includegraphics[width=0.83\textwidth]{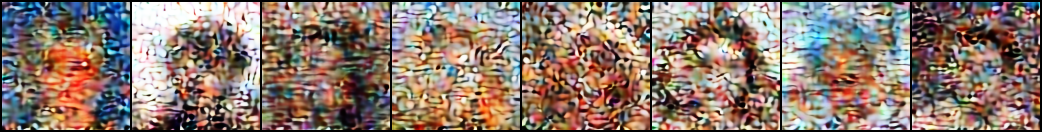}
\end{center}
\vspace{-10pt}  
   \caption{Visual performance on images from the test database. First row: original images | second row: reconstructed ($k=64$) | third row: reconstructed ($k=128$) | forth row: reconstructed from ambiguated codes ($k' = 256$)| fifth row: reconstructed from random guessing of true codes (choosing $k=128$ out of $k'=256$).
   %| sixth row : reconstructing from only $\mathbf{z}^{[1]}$ to $\mathbf{z}^{[8]}$ | seventh row: reconstructing only from $\mathbf{z}^{[12]}$ to $\mathbf{z}^{[20]}$. 
   }
   \label{fig:visual}
   \end{figure*}
%---------------------------------------------------------------------------------------------

%%%.  Sohrab, you removed rows six and seven. Update the caption, please

\vspace{-8pt}

%%%%%%%%% 
\section{Experiments}\label{sec:exp}
\vspace{-5pt}
We conduct experiments on the large-scale CelebA \cite{liu2015faceattributes} database of around $200,000$ images of size $128 \times 128$. We randomly split the dataset and picked $80 \%$ of the images only for training the network, and the rest only for testing. We used the PyTorch \cite{paszke2017automatic} framework to implement and train the above-explained architecture with 6 down-sampling convolutional blocks of ratios $[ 1, 2, 1, 2, 1, 2]$, and a symmetric decoder.  We trained the network for around $40$ epochs using the standard Adam optimizer \cite{kingma2014adam} and settings.

The network was designed to have $L=20$ code-maps, each with length $m=512$, and the sparsity of $k=128$ per item and per code. The storage requirement to describe the support of each item is then $\frac{20 H_2(\frac{128}{512})}{8 \times 1024} \simeq 1.01 \text{KBytes}$, which is very practical for secure communication purposes.

After the training, images from the test set are first encoded and, in order to assess their quality, are then decoded with two sparsity values of $k = 64, 128$. These codes are then ambiguated with random noise to amount for a total sparsity value of $k'=256$ (i.e., $128$ fake components). In order to minimize the chances of the adversary for random guessing, we generate the ambiguation noise from the same distribution of the true non-zero informative components, i.e., the complementary truncated Gaussian distribution with adjusted threshold and variance. Note that even after ambiguation, the public database has a reasonably low storage cost, since the non-zero values can be further quantized without loss of quality. Since the adversary may have the knowledge that the true sparsity was $k=128$, we then try to attack the system by picking $k$ out of $k'$ non-zero values. However, since the distribution of the non-zero activities is highly uniform (due to the network trying to maximize its rate-distortion performance), and the values were also added with the same distribution, this guess can only be random with uniform distribution.

%\textit{Utility \& Privacy Performance:} 
%

Following the \textit{shared secrecy} based on support intersection of data which is introduced by SCA privacy mechanism, the authorized data users can \textit{purify} (unlock) the corresponding ambiguated representation. 
However, the un-authorized parties have no knowledge to `unlock' the public stored representations. 
Fig. \ref{fig:visual} compares the outcomes of different experiments visually and on $8$ randomly-chosen images from the test set. 
We can clearly confirm the high fidelity of the encoded images for the authorized parties, as well as the non-distinguishable quality of reconstruction for the adversary. 
As was expected, the random guessing does not improve the quality, since the chance of zeroing the noise is the same as that of the original content. 
% I am removing the stuff about zeroing out some maps.
%In order to show some of the different attributes learned by the network in each of the code-maps (without ambiguation), we zeroed-out some of the code-maps and reconstruct from partial codes. This is reflected in the last two rows of Fig. \ref{fig:visual}.

%%% I put table at the end of FIRST column of last page, before conclusion
\begin{table}[b!]
\vspace{-0.5cm}
\centering
%\scalebox{0.69}{
\resizebox{\linewidth}{!}{%
\begin{tabular}{l||c|c|c|c|cl}
     & \shortstack{{Recon. ($k=64$)}\\{$\text{rate} = 0.0845$}}  & \shortstack{{Recon. ($k=128$)}\\{$\text{rate} = 0.1690$}}   & \shortstack{{JPEG}\\{$\mathrm{rate} = 0.1830$}}  &  \shortstack{{ambiguated}\\{($k'=256$)}}  & \shortstack{{rand. guess}\\{($k'=256$)}}\\ \hline \hline
     
PSNR & $28.66$  & $30.75$ & $22.40$  & $12.00$ & $12.31$  \\ \hline
SSIM & $0.92$  & $0.95$  &  $0.76$        & $0.24$ & $0.25$ 
\end{tabular}
}
\caption{Rate vs. image quality measures (Results averaged over $200$ randomly-selected images from the test set.) }
\label{table:RD}
\end{table}

As a quantitative comparison and in order to measure the fidelity of the setup for both authorized and public domains, we calculate PSNR (Peak Signal-to-Noise Ratio) and SSIM (Structural Similarity Index) of different experiments in comparison with the original image, as summarized in Table~\ref{table:RD}. As can be seen, the rate-distortion performance of the network is very high, even without lossless entropy coding, and is noticeably surpassing JPEG. This has two reasons, firstly, because of adapting to the content and hence better capturing redundancies than fixed codecs, secondly, the fact that we only encode the support and not the non-zero values. However, these values are not highly entropic and can be quantized, hence surpassing JPEG even by providing privacy utility.

\vspace{-0.3cm}

%%%%%%%%% 
\section{Conclusions} \label{sec:conclusions}
\vspace{-5pt}
We introduced a practical data sharing scheme for privacy-aware image sharing suitable for large-scale setups, where compactness of representations is of important concern for secure communication and storage. This was achieved by ambiguating code-maps of sparse representations, for which we designed a deep CNN as a bottlenecked autoencoder and learned it end-to-end on a large-scale image database.
%We ambiguate a portion of the data with noise and incur combinatorial guessing for the attacker. The key to disambiguate the content is shared only with authorized users, and is designed to be very compact.

\pagebreak
\newpage
\FloatBarrier

% -------------------------------------------------------------------------
\bibliographystyle{IEEEbib}
%\bibliography{strings,refs}
\bibliography{refs}

\end{document}